\documentclass[11pt]{article}
\usepackage{acl2014}
\usepackage{times}
\usepackage{url}
\usepackage{latexsym}

\usepackage{amsmath}
\usepackage{multirow}
\usepackage{graphicx}
\usepackage{booktabs}
\usepackage{graphicx}
\usepackage{epstopdf}
\usepackage{caption}
\usepackage{subcaption}

\title{ Improve the Evaluation of Fluency Using Entropy \\ for Machine Translation Evaluation Metrics }

\author{Hui $\text{Yu}^\dag$\ \ \   Xiaofeng $\text{Wu}^\ddag$ \ \ \ Wenbin $\text{Jiang}^\dag$\ \ \  Qun $\text{Liu}^{\ddag\dag}$\ \ \ Shouxun $\text{Lin}^\dag$\\
 $^\dag$Key Laboratory of Intelligent Information Processing \\
 Institute of Computing Technology, Chinese Academy of Sciences  \\
 $^\ddag$ADAPT Centre, School of Computing, Dublin City University\\
 \\}

\date{}

\begin{document}
\maketitle
\begin{abstract}
The widely-used automatic evaluation metrics cannot adequately reflect the fluency of the translations. The n-gram-based metrics, like BLEU, limit the maximum length of matched fragments to $n$ and cannot catch the matched fragments longer than $n$, so they can only reflect the fluency indirectly. METEOR, which is not limited by n-gram, uses the number of matched chunks but it does not consider the length of each chunk.  
In this paper, we propose an entropy-based method, which can sufficiently reflect the fluency of translations  through the distribution of matched words. This method can easily combine with the widely-used automatic evaluation metrics to improve the evaluation of fluency. 
Experiments show that the correlations of BLEU and METEOR are improved on sentence level after combining with the entropy-based method on WMT 2010 and WMT 2012.

\end{abstract}
\section{Introduction}
Automatic machine translation (MT) evaluation plays an important role in the evolution of MT. It not only evaluates the performance of MT systems, but also provides guidance for the improvement of MT systems \cite{och2003minimum}. 

The automatic MT evaluation metrics can be classified into three categories: lexicon-based methods \cite{papineni2002bleu,Snover06astudy,Lavie:2007:MAM:1626355.1626389,chen-kuhn:2011:WMT,Chen:2012:IAM:2393015.2393021}, syntax-based methods \cite{liu2005syntactic,Owczarzak:2007:LDM:1626355.1626369,Chan08maxsim:a,Zhu:2010:SPS:1944566.1944741,mdobrink:BLEUATRE} and semantic-based methods \cite{lo-tumuluru-wu:2012:WMT}, according to the employed information type. 
Most of the lexicon-based metrics  obtain the similarity between the reference and hypothesis based on n-gram, such as BLEU \cite{papineni2002bleu} and  NIST\cite{doddington2002automatic}. 
BLEU 
obtains the score by a geometric mean of the n-gram precisions and a length-based penalty. NIST is closely related with BLUE but uses the arithmetic mean instead of geometric mean.  
For these metrics, the maximum length of matched fragments is limited to $n$, so they cannot catch the matched fragments longer than $n$. 
Some metrics which are not limited by n-grams relieve this problem, such as METEOR \cite{Lavie:2007:MAM:1626355.1626389}. METEOR uses the Fmean of unigrams and a penalty. 
The penalty in METEOR is related to the number of matched chunks\footnote{The words in each chunk are in adjacent positions in the hypothesis, and are also mapped to unigrams that are in adjacent positions in the reference.}. When the number of chunks in two sentence are the same, 
METEOR doesn't distinct them.
The syntax-based metrics obtain the similarity by comparing the syntactic structures of two trees, and they 
 cannot reflect the fluency directly. Semantic-based metrics, such as MEANT \cite{lo-tumuluru-wu:2012:WMT}  which uses semantic role labeling (SRL) to match the predicate and arguments, mainly obtain the semantic information and do not consider the fluency.

In this paper, we propose an entropy-based method which can not only exploit the chunks with the maximum matched length but also reflect the difference between the lengths of the chunks. 
This method can easily combine with the widely-used automatic evaluation metrics to improve the evaluation of fluency. In the experiments, the new method is used to combine with BLEU and METEOR, and the sentence level correlations of BLEU and METEOR are improved on WMT 2010 and WMT 2012.

\section{Entropy-based Method}

 
In this section, we introduce entropy and the entropy-based method (ENT) which can reflect the fluency of translations.
\subsection{Entropy}
Entropy is a measure of the uncertainty in a random variable. Shannon denoted the entropy H of a discrete random variable x with possible values ${x_1,x_2,...,x_n}$. The entropy is defined as Formula \eqref{entropy} \cite{shannon2001mathematical}.
\begin{equation}
H(X)= -\sum_{i=1}^{n} P(x_i) log_2 P(x_i)
\label{entropy}
\end{equation}
 $P(x_i)$ is the probability of $x_i$ showing up in the stream of characters. 
The more decentralized of the values ${x_1,x_2,...,x_n}$, the higher of the entropy H(X). So the entropy can reflect the distribution of the values of variable x. 

\subsection{Entropy-based Method}
  
In the automatic evaluation of machine translation, entropy can reflect the distribution of matched words. A lower entropy corresponds to a more concentrate distribution of matched words which represents a more fluent hypothesis. On the contrary, a higher entropy corresponds to a more disperse distribution of matched words, which represents a less fluent hypothesis.
So the entropy-based method can reflect the fluency of translations sufficiently by the distribution of the words.

An example (a reference and three hypotheses) is shown as follows.

\begin{itemize}
\item
ref:\ \ There are books on the desk
\item
hyp1:\ \textbf{There are books} in that \textbf{desk}
\item
hyp2:\ \textbf{There are} table  \textbf{on the} book
\item
hyp3:\ \textbf{There are} table  \textbf{on } book \textbf{the}
\end{itemize}
The matched words are in bold. hyp1, hyp2 and hyp3 can all match four words, but the distribution of the four words are different. The matched words are in two chunks for hyp1 and hyp2, and three chunks for hyp3. A smaller number of chunks represents a more concentrated distribution of the matched words, and corresponds to a more fluent hypothesis. From this point of view, hyp1 and hyp2 are better than hyp3. hyp1 has the same number of chunks as hyp2 but the number of the matched words in the two chunks is (3, 1) for hyp1 and (2, 2) for hyp2. hyp1 is considered to be more fluent than hyp2.   
 
The details of the ENT are represented in following three steps. 
First, we obtain the matched words through the alignment of reference and hypothesis. The alignment is derived using Meteor Aligner\footnote{\url{http://www.cs.cmu.edu/~alavie/METEOR/}}. 
The matched words are considered to be in a chunk if they are continuous and appear in the same order in both reference and hypothesis. 
Second, the entropy of chunks is calculated using Formula \eqref{entropy-mt}.
\begin{equation}
H=-\sum_{i=1}^c\dfrac{l_i}{L}log(\dfrac{l_i}{L})
\label{entropy-mt}
\end{equation}
 $l_i$ is the length of the $ith$ chunk. $c$ is the number of the chunks. $L$ is the total number of the matched words. In the last step, the final score of ENT is achieved by Formula \eqref{ents}. To obtain a score within scope (0,1), an exponential function is used. We use $-H$ instead of $H$ in the formula to ensure that a higher score of ENT represents a more fluent translation.
\begin{equation}
ENT=\alpha ^{-H \times LP}, \ \ \  \alpha \in (1,1.5)
\label{ents}
\end{equation} 
$LP$, a length penalty, is calculated by Formula \eqref{len-pen}. $l_h$ is the length of hypothesis. $l_r$ is the length of reference.
\begin{equation}
 LP=\beta^{|{\dfrac{l_h}{l_r}-1}|}, \ \ \ \beta \in (1,2)
 \label{len-pen}
\end{equation}

Using Formula \eqref{ents}, the scores in the above example can be obtained as follows. 
\begin{equation*}
LP_{hyp1}=LP_{hyp2}=LP_{hyp3}=\beta^{|\frac{6}{6}-1|}=1
\end{equation*}
\begin{equation*}
ENT_{hyp1} = \alpha^{-(-(\frac{3}{4}log\frac{3}{4} + \frac{1}{4}log\frac{1}{4})) \times 1 } \approx \alpha^{-0.24}
\label{exa-1}
\end{equation*}
\begin{equation*}
ENT_{hyp2} = \alpha^{-(-(\frac{2}{4}log\frac{2}{4} + \frac{2}{4}log\frac{2}{4})) \times 1} \approx \alpha^{-0.30}
\label{exa-2}
\end{equation*}
\begin{equation*}
ENT_{hyp3} = \alpha^{-(-(\frac{2}{4}log\frac{2}{4} + 2 \times \frac{1}{4}log\frac{1}{4} )) \times 1} \approx \alpha^{-0.45}
\label{exa-3}
\end{equation*}

We can see that $ENT_{hyp1} > ENT_{hyp2} > ENT_{hyp3} $. Accordingly, the quality of hyp1 is obviously better than hyp2, and hyp2 is better than hyp3. So the entropy-based method can distinct these situations well.

The alignment of reference and hypothesis is derived only using \textit{exact} match for ENT. We can also use linguistic resources to get the alignment, such as stem\cite{porter2001snowball}, synonym (Wordnet\footnote{\url{http://wordnet.princeton.edu/}}) and paraphrase. In this case, we name the new method as ENTplus (ENTp).

\section{Combine Entropy-based Method with Other Metrics}
The new entropy-based method can effectively measure the fluency of a sentence. 
Most of the current metrics are good at the measure of accuracy, so we combine the entropy-based method with the widely-used automatic evaluation metrics to further improve the performance of these metrics. In this section, we take BLEU and METEOR as examples to introduce the combination method, but the entropy-based method can combine with most of the widely-used evaluation metrics. 

\subsection{Combine Entropy-based Method with BLEU}
BLEU is a widely-used automatic evaluation metric owing to its simplicity and effectiveness. BLEU is calculated by Formula \eqref{bleu} \cite{papineni2002bleu}.

\begin{equation}
BLEU=exp(\sum_{n=1}^N w_nlogp_n) \times BP
\label{bleu}
\end{equation}

\begin{equation}
BP=\left\{
\begin{aligned}
 1&  & if\ c>r \\
e^{(1-r/c)}&  & if\ c\leq r
\end{aligned}
\right.
\label{BP}
\end{equation}
In Formula \eqref{bleu}, the first part is a geometric mean of the n-grams precision where $p_n$ is the precision of $n$-gram, and the second part is a length-based penalty as shown in Formula \eqref{BP}.
There is also a length penalty in ENT. So we remove the part of length penalty in ENT when combining ENT with BLEU (Formula \ref{bleu-ent}). The experience value of $\alpha$ is 1.05.
\begin{equation}
\text{BLEU}_{ENT}=BP\times exp(\sum_{n=1}^N w_nlogp_n) \times \alpha^{-H}
\label{bleu-ent}
\end{equation}

\subsection{Combine Entropy-based Method with METEOR}
METEOR is calculated by Formula \eqref{meteor}, in which $Pen$ is calculated by Formula \eqref{Pen}. 
\begin{equation}
\text{METEOR}=Fmean \times (1-Pen)
\label{meteor}
\end{equation}
 
 \begin{equation}
 Pen=x1 \cdot (\dfrac{\#chunks}{\#unigrams\_matched})^{x2}
 \label{Pen}
 \end{equation}
The first part in Formula \eqref{meteor} is the fmean of unigrams. The second part is related with the number of chunks. METEOR doesn't consider the length of each chunk, so it cannot reflect the situation that two hypotheses have the same number of matched unigrams and the same number of chunks, but different lengths for each chunk. We use ENT instead of $1-Pen$ to reflect the above situation, and the final score can be computed in Formula \eqref{meteor-ent}. The experience value of $ \alpha $ and $ \beta $ are 1.5 and 1.12 respectively.
\begin{equation}
\text{METEOR}_{ENT} = Fmean \times \alpha^{-H\times LP}
\label{meteor-ent}
\end{equation}

\section{Experiments}	

To compute the correlation with human judges on sentence level, Kendall's rank correlation coefficient $\tau$ is employed. A higher value of $\tau$ means a better ranking similarity with the human judgments. $\tau$ is calculated as follows.

\begin{equation*}
\tau = \dfrac{count_{con\ pairs} - count_{dis\ pairs}}{count_{total\ pairs}}
\end{equation*}
$ count_{con\ pairs} $ is the count of concordant pairs. $ count_{dis\ pairs} $ is the count of discordant pairs.

\subsection{Data}

In order to verify the effectiveness of ENT, we carry out the experiments on WMT 2010 and WMT 2012.
There are four language pairs including German-to-English (de-en), Czech-to-English (cz-en), French-to-English (fr-en) and Spanish-to-English (es-en), which are all derived from WMT 2010 with 2034 sentences and WMT 2012 with 3003 sentences. 
The number of translation systems for each language pair is showed in Table \ref{data}. 

\begin{table}[ht]
\centering
\begin{tabular}{|l|c|c|c|c|}
\hline
data&cz-en &de-en& es-en &fr-en \\
\hline
WMT2010& 12&25 &15 &24 \\
\hline
WMT2012 &6 &16 & 12& 15\\
\hline
\end{tabular}
\caption{The number of translation systems for each language pair on WMT 2010 and WMT 2012.
}
\label{data}
\end{table}

\begin{table*}[!htb]
 \centering
\begin{tabular}{|l|l|c|c|c|c|l|}
\hline
Data & Metrics  & cz-en & de-en & es-en & fr-en & ave\\
\hline
\multicolumn{1}{|l|}{\multirow{3}{*}{WMT10}}&BLEU &0.2554 &0.2748 &0.2805 &0.2197 & 0.2576\\
\cline{2-7}
&BLEU+ENT &0.2565 &0.2730 &0.2822 &0.2211 & 0.2582 \\
\cline{2-7}
&BLEU+ENTp &\textbf{0.2643} &\textbf{0.2823} &\textbf{0.3010} & \textbf{0.2368}& \textbf{0.2711}(+1.35)\\
\hline
\multicolumn{1}{|l|}{\multirow{3}{*}{WMT12}}&BLEU & 0.1567 &0.1840 &0.1938 & 0.1999 & 0.1836 \\
\cline{2-7}
&BLEU+ENT &0.1660 &0.1907 & 0.1940& 0.2060 & 0.1892 \\
\cline{2-7}
&BLEU+ENTp& \textbf{0.1732}& \textbf{0.1989 }&\textbf{0.2052} &\textbf{0.2208} & \textbf{0.1995}(+1.59) \\
\hline
\end{tabular}
\caption{Sentence level correlations of BLEU, BLEU+ENT and BLEU+ENTp on WMT 2010 and WMT 2012. The last column gives the average scores of the four language pairs. 
}
\label{rst-bleu}
\end{table*}

\begin{table*}[!htb]
 \centering
\begin{tabular}{|l|l|c|c|c|c|l|}
\hline
Data & Metrics  & cz-en & de-en & es-en & fr-en & ave\\
\hline
\multicolumn{1}{|l|}{\multirow{2}{*}{WMT10}}&METEOR &0.3292 &0.3585 & 0.3283& 0.2710& 0.3218\\
\cline{2-7}
&METEOR+ENTp &\textbf{0.3354} &\textbf{0.3593} &\textbf{0.3586} &\textbf{0.2923} &\textbf{0.3364}(+1.46)\\
\hline
\multicolumn{1}{|l|}{\multirow{2}{*}{WMT12}}&METEOR &0.2124 &0.2748 &0.2493 &0.2506 & 0.2468\\
\cline{2-7}
&METEOR+ENTp & \textbf{0.2153}& 0.2730 &\textbf{0.2585} & \textbf{0.2539}& \textbf{0.2502}(+0.34) \\
\hline
\end{tabular}
\caption{Sentence level correlations of METEOR and METEOR+ENTp on WMT 2010 and WMT 2012. The last column gives the average scores of the four language pairs. 
}
\label{rst-meteor}
\end{table*}

\subsection{Experiment Results}

The correlations of BLEU\footnote{ftp://jaguar.ncsl.nist.gov/mt/resources/mteval-v13a.pl}  are the results of 4-gram with smoothing option. 
According to the different methods of obtaining the chunks, we try two configurations, BLEU+ENT and BLEU+ENTp. BLEU+ENT is only using the exact match. BLEU+ENTp is using some resources which are stem, synonym and paraphrase. 
The correlations of METEOR are obtained from the released data of WMT 2010 (Version 1.27 ) and WMT 2012 (Version 1.48 ) with task option ‘rank’. 
We only do the experiment using outside resources (METEOR+ENTp), because METEOR also uses the outside resources. \footnote{Interested readers can find the source code of ENT and ENTp from https://github.com/YuHui0117/AMTE/tree/master/ENTFp.}

The sentence level correlations of the four language pairs and the average scores  are shown in Table \ref{rst-bleu} and Table \ref{rst-meteor}. 
In Table \ref{rst-bleu}, BLEU+ENT is better than BLEU on both WMT10 and WMT12, but the result is only improved a little compared with BLEU. The reason is that the alignment is not good enough when only using exact match. When using stem, synonym and paraphrase, the result has a significant improvement of 1.35 on WMT 2010 and 1.59 on WMT 2012 respectively when comparing with BLEU. The number of reference is limited, and linguistic resources can enrich the reference, so ENTp can get better performance than ENT.

From Table \ref{rst-meteor}, we can see that METEOR+ENTp has a significant improvement (1.46 on average) on WMT 2010, while  the improvement on WMT 2012 (0.34 on average) is not as much as on WMT 2010. The METEOR version on WMT 2012 optimizes the parameters on the data of WMT 2009 and WMT 2010. We didn't tune the parameters after combing METEOR with entropy-base method, so the improvement is not very significant. 

In all,  when combining the entropy-based penalty with the widely-used automatic evaluation metrics, such as BLEU and METEOR, the performance can be improved, which proves the effectiveness of the entropy-based method.

\section{Conclusion and Future Work}
In this paper, we use entropy to reflect the fluency of the translation, and propose an  entropy-based method. When combing the entropy-based method with the widely-used automatic evaluation metrics, such as BLEU and METEOR, the performances of these metrics are improved.

One purpose of automatic evaluation metrics is to improve the  quality of machine translation systems. So, in the future, we will use the combination of entropy-based method and widely-used metrics in the tuning process to improve the translation quality, such as MERT (Minimum Error Rate Training) \cite{och2003minimum}. 

\section*{Acknowledgements}
This work is supported by National Natural Science Foundation of P.
R. China under Grant Nos. 61379086, 61602284, 61602285, 61602282 and
Shandong Provincial Natural Science Foundation of China under Grant
No. ZR2015FQ009. Qun Liu's work is partially supported by the
Science Foundation Ireland (Grant 13/RC/2106) as part of the
ADAPT Centre at Dublin City University.



\bibliographystyle{acl}
\bibliography{hui}

\end{document}